\pdfoutput=1
\documentclass[11pt]{article}

\usepackage{EMNLP2022}

\usepackage{times}
\usepackage{latexsym}

\usepackage[T1]{fontenc}
\usepackage[utf8]{inputenc}

\usepackage{microtype}

\usepackage{graphicx}
\usepackage{subfigure}
\usepackage{multirow}

\usepackage{amsmath}
\usepackage{mathtools}

\usepackage{enumitem}
\setenumerate[1]{itemsep=2pt,partopsep=0pt,parsep=\parskip,topsep=5pt}
\setitemize[1]{itemsep=2pt,partopsep=0pt,parsep=\parskip,topsep=5pt}
\setdescription{itemsep=2pt,partopsep=0pt,parsep=\parskip,topsep=5pt}

\usepackage{ntheorem}
\theoremseparator{:}

\usepackage{inconsolata}

\title{ACL: \underline{A}ligned \underline{C}ontrastive \underline{L}earning Improves BERT and Multi-exit BERT Fine-tuning  }

\author{
    Liz Li\textsuperscript{1,}, Wei Zhu\textsuperscript{2,}\thanks{\ \ Corresponding author. For any inquiries, please contact: michaelwzhu91@gmail.com. }\\
    \small \textsuperscript{1}\ DataSelect AI, Xuhui, Shanghai, China \\
    \small \textsuperscript{2}University of Hong Kong, Hong Kong, HK, China
}

\begin{document}
\maketitle
\begin{abstract}

Despite its success in self-supervised learning, contrastive learning is less studied in the supervised setting. In this work, we first use a set of pilot experiments to show that in the supervised setting, the cross-entropy loss objective (CE) and the contrastive learning objective often conflict with each other, thus hindering the applications of CL in supervised settings. To resolve this problem, we introduce a novel \underline{A}ligned \underline{C}ontrastive \underline{L}earning (ACL) framework. First, ACL-Embed regards label embeddings as extra augmented samples with different labels and employs contrastive learning to align the label embeddings with its samples' representations. Second, to facilitate the optimization of ACL-Embed objective combined with the CE loss, we propose ACL-Grad, which will discard the ACL-Embed term if the two objectives are in conflict. To further enhance the performances of intermediate exits of multi-exit BERT, we further propose cross-layer ACL (ACL-CL), which is to ask the teacher exit to guide the optimization of student shallow exits. Extensive experiments on the GLUE benchmark results in the following takeaways: (a) ACL-BRT outperforms or performs comparably with CE and CE+SCL on the GLUE tasks; (b) ACL, especially CL-ACL, significantly surpasses the baseline methods on the fine-tuning of multi-exit BERT, thus providing better quality-speed tradeoffs for low-latency applications.\footnote{Codes will be made publicly available upon acceptance. }

% dual contrastive learning (DualCL) framework that simultaneously learns the features of input samples and the parameters of classifiers in the same space. Specifically, DualCL regards the parameters of the classifiers as augmented samples associating to different labels and then exploits the contrastive learning between the input samples and the augmented samples. Empirical studies on five benchmark text classification datasets and their lowresource version demonstrate the improvement in classification accuracy and confirm the capability of learning discriminative representations of DualCL.

\end{abstract}

\section{Introduction}

Recently, contrastive learning has been proved repeatedly to be effective in unsupervised learning to obtain generic representations for downstream supervised tasks~\cite{Wu2018UnsupervisedFL,Oord2018RepresentationLW,Tian2020ContrastiveMC,He2020MomentumCF,Chen2021ExploringSS,Cui2023UltraFeedbackBL,wang2024ts,yue2023-TCMEB,Zhang2023LearnedAA,2023arXiv230318223Z,Xu2023ParameterEfficientFM,Ding2022DeltaTA,Xin2024ParameterEfficientFF,qin2023chatgpt,PromptCBLUE,text2dt_shared_task,Text2dt,zhu_etal_2021_paht,Li2023UnifiedDR,Zhu2023BADGESU,Zhang2023LECOIE,Zhu2023OverviewOT,guo-etal-2021-global,zhu-etal-2021-discovering,Zheng2023CandidateSF,info:doi/10.2196/17653,Zhang2023NAGNERAU,Zhang2023FastNERSU,Wang2023MultitaskEL,Zhu2019TheDS,zhu2021leebert,Zhang2021AutomaticSN,Wang2020MiningIH,li2025ft,leong2025amas,zhang2025time,yin2024machine,zhu2026mrag,zhu2026evaluatechatgpt}, by asking the model to represent different views of a sample as close as possible and represent different samples as distinct as possible. Despite the majority of literature on contrastive learning works in the unsupervised learning setting, supervised contrastive learning has also been adapted to the supervised settings~\cite{Khosla2020SupervisedCL}. \citet{Khosla2020SupervisedCL} construct the supervised contrastive learning (SCL) objective to ask the representations of examples in the same class to be close and those from different classes to be distant. The idea of SCL has also been developed in the field of NLP~\cite{Gunel2021SupervisedCL,tian2024opportunities,hersh2024search,tian2024opportunities,hersh2024search,zhu2024iapt,zhu-tan-2023-spt,Liu2022FewShotPF,xie2024pedro,Cui2023UltraFeedbackBL,zheng2024nat4at,zhu2023acf,gao2023f,zuo-etal-2022-continually,zhang-etal-2022-pcee,sun-etal-2022-simple,zhu-etal-2021-gaml,Zhu2021MVPBERTMP,li-etal-2019-pingan,zhu2019panlp,zhu2019dr,zhou2019analysis,zhang2025time,wang2025ts,liu2025parameter,yi2024drum,tian2024fanlora}. However, despite the good results, we argue that the current SCL approach still has room for improvements. One intuitive reason is that the SCL objective may not align with the CE objective, thus resulting in poor optimization.

%  Contrastive learning is effective because it can achieve both “alignment” and “uniformity”~\cite{Wang2020UnderstandingCR,Gao2021SimCSESC}. Emerging from computer vision, Contrastive learning has been widely adopted in other fields like time series, graph neural networks and natural language process (NLP) \cite{Jaiswal2020ASO}. 

% Note that despite the majority of literature on contrastive learning works on the unsupervised learning setting, supervised contrastive learning has also been adapted to the supervised settings~\cite{Khosla2020SupervisedCL}. \citet{Khosla2020SupervisedCL} construct the supervised contrastive learning (SCL) objective to ask the representations of examples in the same class to be close and those from different classes to be distant. The idea of SCL has also been developed in the field of NLP~\cite{Gunel2021SupervisedCL}. \citet{Gunel2021SupervisedCL} shows that combine the SCL objective with the cross-entropy (CE) objective (CE+SCL) is beneficial for downstream tasks, both in performance scores and robustness.  

\begin{figure*}[h]
	\centering
		\includegraphics[width=0.95\textwidth]{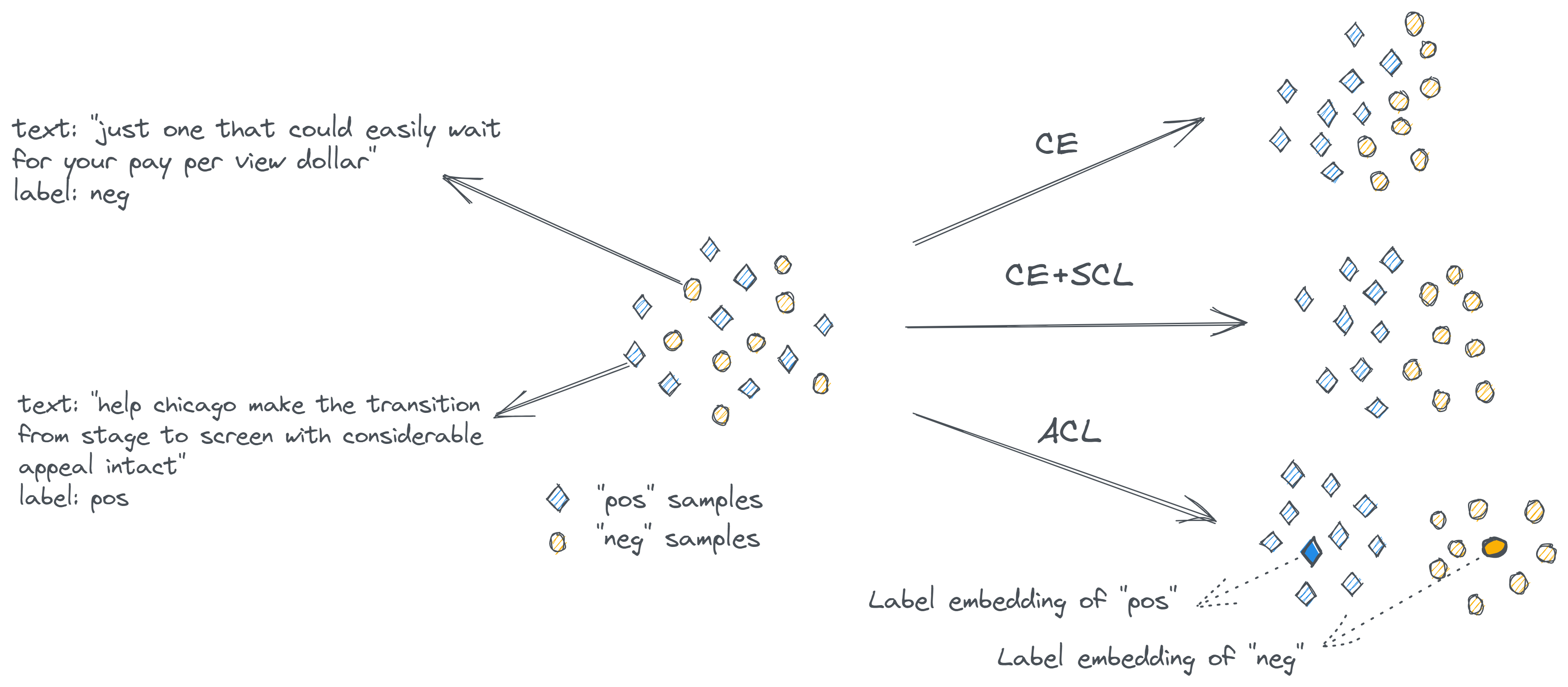}
		\label{subfig:framewrok}
	\caption{\label{fig:framework} Our proposed ACL framework. ACL differs from the vanilla SCL objective in that it also aligns the semantic representations of input samples with the label embeddings through the contrastive learning objective. Intuitively, ACL considers the label embeddings as anchors to encourage more apparent separations of samples of different labels. Although we show a binary classification case for simplicity, the loss is generally applicable to any multi-class classification setting. 
 }
\end{figure*}

% when it learns to push examples from the same label close and examples from different classes further apart, 

% We show examples from the SST-2 sentiment analysis dataset from the GLUE benchmark, where label "neg" (shown in red) is negative movie reviews and label "pos" (shown in blue) is positive movie reviews.

In this work, we first conduct a pilot experiment to show that: the gradients of SCL and CE have quite different directions, even after being jointly optimized (see Figure~\ref{fig:gradient_angle}). To overcome this issue, we propose a novel contrastive learning framework, \underline{A}ligned \underline{C}ontrastive \underline{L}earning (ACL) (depicted in Figure~\ref{fig:framework}). Our ACL framework consists of two parts: (a) ACL-Embed, which includes the learnable label embeddings from the classifier in the contrastive learning objective. ACL-Embed uses label embeddings as anchors to separate the representations from different classes. (b) ACL-Grad, which discards the ACL-Embed loss term if its gradient has a quite different direction (or angle > 90$^{\circ}$) than the one from CE loss. ACL-Grad makes the loss function adjustments adaptively on a given batch of samples. 

We further investigate whether our ACL framework can help to improve the multi-exit BERT fine-tuning. Multi-exit BERT installs an intermediate classifier (or exit) on each layer of BERT. Multi-exit BERT is the backbone for early exiting, a vital model inference speedup method~\cite{Teerapittayanon2016BranchyNetFI,Kaya2019ShallowDeepNU,Zhou2020PABEE,Zhu2021LeeBERTLE,Zhu2021GAMLBERTIB}. Early exiting can adaptively decide how many layers of BERT the sample should go through before making the final prediction using one of the intermediate exits. Due to its ability to flexibly adjust the average latency, early exiting has attracted attention from academia and industry. 

To help improve the overall performances of multi-exit BERT, we propose a cross-layer ACL objective (ACL-CL), that is, to treat the last layer as the teacher and incorporate the teacher's sample representations and label embeddings in the intermediate exit's ACL objective. Intuitively, we use the ACL objective as a knowledge distillation approach so that the stronger teacher exit can guide the optimization direction of the weaker student exit.

We conduct extensive experiments and ablation studies on the GLUE benchmark~\cite{Wang2018GLUEAM} with BERT~\cite{devlin-etal-2019-bert} and RoBERTa~\cite{Liu2019RoBERTaAR} backbones. For BERT/RoBERTa fine-tuning, we show that our ACL objective achieves performance improvements over vanilla fine-tuning or CE+SCL approach~\cite{Gunel2021SupervisedCL} on 5 of the seven classification tasks. Our ACL framework is quite effective in multi-exit BERT/RoBERTa fine-tuning. With the RoBERTa backbone, our ACL-CL objective can outperform TinyBERT/DistillBERT at the 6-th layer without further pre-training. 

Our contributions are three-fold: 
\begin{itemize}
    \item We propose ACL, a novel contrastive learning framework in supervised settings. ACL consists of two components, ACL-Embed and ACL-Grad. 
    \item We propose the cross-layer ACL objective for multi-exit BERT fine-tuning, an effective way to distill knowledge from the deeper exits.
    \item Experiments on the GLUE benchmark demonstrate the effectiveness of our ACL framework in BERT fine-tuning and multi-exit BERT fine-tuning. 
    
\end{itemize}

\section{Related Work}

\subsection{Contrastive learning}

\noindent \textbf{Contrastive learning in the supervised setting}\footnote{See Appendix \ref{sec:appendix_additional_related_works} for related works on contrastive learning in the unsupervised settings} \quad Contrastive learning has been extended to the supervised setting. \citet{Khosla2020SupervisedCL} propose the supervised contrastive loss (SCL) that pulls the representations of images from the same class closer while pushing those from different classes further. \citet{Gunel2021SupervisedCL} also develops a version of SCL in the context of BERT fine-tuning.

Our work complements this branch of literature by extending \citet{Gunel2021SupervisedCL} to propose a novel supervised contrastive learning objective, ACL, which includes the label embedding from the classifiers in the contrastive objectives. In addition, we propose stabilizing the CL and CE objective joint optimization adaptively based on gradient directions.

\subsection{Multi-Exit BERT training}

\noindent Adaptive inference method like early exiting~\cite{Teerapittayanon2016BranchyNetFI,Kaya2019ShallowDeepNU,Zhou2020PABEE,Zhu2021LeeBERTLE,Zhu2021GAMLBERTIB} is an important branch of literature focusing on applying different parts of BERT to process different samples, achieving inference speedup on average. It is in parallel with and can work together with static model compression methods~\cite{Tambe2020EdgeBERTOO} like network pruning \cite{zhu2017prune,xu2020bert,fan2019reducing,gordon2020compressing}, student network distillation \cite{sun2019patient,sanh2019distilbert,jiao2019tinybert}. 

Early exiting requires a multi-exit BERT, a BERT backbone with an intermediate classifier (or exit) installed on each layer. Early exiting literature mainly focuses on the design of the early exiting strategies, that is, determining when an intermediate exit's prediction is suitable as the final model prediction. Score based strategies~\cite{Teerapittayanon2016BranchyNetFI,xin2020deebert,Kaya2019ShallowDeepNU,xin2021berxit}, prior based strategies~\cite{sun-etal-2022-simple} and patience based strategies~\cite{Zhou2020PABEE} have been proposed. However, the literature focuses less on the training of the multi-exit neural network, which is a major factor for early exiting performances. There are three types of training methods for training the multi-exit neural network: (a) joint training (JT)~\cite{Teerapittayanon2016BranchyNetFI}, that is, all the exits are jointed optimized together with the fine-tuning of BERT. (b) two-stage training method (2ST)~\cite{liu2020fastbert,xin2020deebert}, which first fine-tune the backbone BERT and the last layer's exit till convergence, and then at the second stage, the backbone and the last exit are fixed, and the intermediate exit is trained by distilling knowledge from the last exit. (c) Berxit~\cite{xin2021berxit} propose an alternating training (ALT) method, combining 2ST and JT. 

Our work complements the literature on Early exiting or multi-exit neural networks by applying cross-layer contrastive learning (our ACL-CL approach) to exploit the potentials of shallow BERT layers better and improve the overall performances of multi-exit BERT.

\section{Preliminaries}

\noindent In this section, we introduce the necessary background for BERT early exiting. we consider the case of multi-class classification with $K$ classes, $\mathcal{K} = \left\{1, 2, ..., K\right\}$. During training, the model will receive a batch $\left\{(x_i, y_i), i \in \mathcal{I} = \left\{1, 2, ..., N\right\} \right\}$ containing $N$ samples, where $x_i \in \mathbf{R}^{L}$ is the input sentence consisting of $L$ words, and $y_i \in \mathcal{K}$ is the label.

% \subsection{Backbone Models}

% \noindent In this work, we adopt BERT as backbone model. BERT is a multi-layer transformer network that has been pre-trained on a huge corpus in a self-supervised way.

\subsection{Contrastive learning in the supervised setting}

\citet{Khosla2020SupervisedCL} and \citet{Gunel2021SupervisedCL} incorporates samples' label information to CL in a straightforward way.\footnote{Reader can refer to Appendix~\ref{subsec:appendix_preliminaries_cl} for a brief introduction to the contrastive learning objective for the unsupervised setting. } It simply takes the inputs with the same label in the batch as positive samples and those from different classes as negative samples. Let $\Phi(\cdot) \in \mathbf{R}^{d}$ denotes an encoder that outputs the $l_{2}$ normalized final representation; $N_{y_i}$ is the total number of examples in the batch that
have the same label as $y_i$; $\tau$ is the scalar temperature parameter that controls the
separation of classes. The supervised contrastive objective is:
\begin{flalign}
s(i, j) & = \dfrac{ \exp (\Phi(x_i) \cdot \Phi(x_j) / \tau ) }{ \sum_{k=1}^{N} \exp (\Phi(x_i) \cdot \Phi(x_k) / \tau)  }, \nonumber \\
\mathcal{L}_{sup} & = \sum_{i=1}^{N} \dfrac{-1}{N_{y_i} - 1} \sum_{j=1}^{N} \mathbf{1}_{i\neq j} \mathbf{1}_{y_i = y_j} \log s(i, j). \nonumber \\
\label{eq:scl}
\end{flalign}
Let $\lambda$ denote a scalar weighting hyperparameter, then the overall loss is a weighed average of the CE loss (denoted as $\mathcal{L}_{ce}$) and the SCL loss: $ \mathcal{L} = (1-\lambda) \mathcal{L}_{ce} + \lambda \mathcal{L}_{sup}$. 

Supervised contrastive learning motivates to learn the generic semantic representations that help separate the samples into different classes. However, we are asking whether the optimization of CL and CE loss terms will always agree with one another? In this work, we will look closely into this aspect of SCL.

\subsection{Early Exiting} 

\noindent Early exiting is based on multi-exit BERT, which are BERT with classifiers (or exits) at each layer. With $M$ layers, $M$ classifiers $f_m(x;\theta_m):\mathcal{X}\rightarrow \Delta^K (m=1,2,...M)$ are designated at M layers of BERT, each of which maps its input to the probability simplex $\Delta^K$. $f_m(x;\theta_m)$ can take the form of a simple linear layer (linear exits) following~\cite{Zhou2020PABEE}. 

However, in this work, we find that adding a multi-head self-attention layer to the exit (MHA exit) can boost the performances of intermediate layers. Representations of the current BERT layer, $H^{(m)}_{i}$, will first be down-sampled to a smaller dimension $d_e$ (e.g., 64). Then, it will go through a Tanh activation, followed by a multi-head self-attention operation~\cite{Vaswani2017AttentionIA}. Then the [CLS] token's vector representation will be passed to the second Tanh activation, and the final representation in the exit is denoted as $h^{e}_{i}$. $h^{e}_{i}$ will be fed into a linear layer to generate the predicted logits. This work considers the MHA exit as the default exit architecture. 

% , and an cell for pooling layer, and finally another activation cell.

% Denote the BERT encoded representation of sample $i$ as $H_i \in \mathbf{R}^{L \times d}$. A MHA exit first project $H_i$ into a lower dimension $d_e$ to $H_{i}^{(1)}$ via a linear layer:
% \begin{equation}
%     H_{i}^{(1)} = % \mathrm{Tanh}(H_{i}^{(1)}W_1 + b_1),
% \end{equation}
% \noindent where $Tanh(\cdot)$ is the hyperbolic activation function. Then a standard multi-head self-attention operation~\cite{Vaswani2017AttentionIA} will further encode $H_{i}^{(1)}$ to $H_{i}^{(2)}$ without dimensionality changes. Then the vector representation $h^{(2)}_{i, 0} \in \mathbf{R}^{d_e}$ of the starting token $[CLS]$ is regarded as the representation for the sentence. Then a linear layer will give out the exit's final representation $h^{e}_{i} \in \mathbf{R}^{d_e}$:
% \begin{equation}
%     h^{e}_{i} = \mathrm{Tanh}(h^{(2)}_{i, 0} W_2 + b_2).
% \end{equation}
% \noindent $ 

\subsection{Multi-exit BERT training}
\label{sec:multi_exit_bert_training}

Due to the length limit, readers are referred to Appendix \ref{subsec:mutli_exit_training} for other multi-exit BERT training methods. Now, we introduce the 2ST training method.

\textbf{2ST}. The two-stage (2ST) \cite{xin2020deebert,liu2020fastbert} training strategy\footnote{Due to length limit, reader are referred to Appendix \ref{subsec:mutli_exit_training} for other multi-exit BERT training methods.} divides the training procedure into two stages. The first stage is identical to the vanilla BERT fine-tuning, updating the backbone model and only the final exit. In the second stage, we freeze all parameters updated in the first stage and fine-tune the remaining exits separately:
\begin{flalign}
    &\mathrm{Stage 1}: \mathcal{L}_{stage1} = \mathcal{L}^{CE}_{M}(y_i,f_M(x_i;\theta_M)) \nonumber    \\
    &\mathrm{Stage 2}: \mathcal{L}_{stage2}=\mathcal{L}_m^{CE}, m = 1, ..., M-1. \label{training:2st}  \nonumber
\end{flalign}
\noindent where $\mathcal{L}^{CE}_{m}=\mathcal{L}_m^{CE}(y_i,f_m(x_i;\theta_m))$ denotes the cross-entropy loss of $m$-th exit. The 2ST training's performances can be further enhanced by treating the last exit trained in the first stage as the teacher model and conducting knowledge ditillation~\cite{Hinton2015DistillingTK}, as is demonstrated by \citet{liu2020fastbert}.

\section{Aligned Contrastive Learning}

In this section, we elaborate on our ACL framework. The core idea of ACL is to make the contrastive learning objective in agreement with the learning of the classifier.

\subsection{Motivation}

Combining an SCL objective with the CE loss could help the BERT backbone model extract generic semantic representations. However, through our preliminary experiments, we observe that the weight of the SCL loss $\lambda$ has a significant impact on the model performances, and the better performances are achieved by setting $\lambda$ to be relatively small (e.g., 0.02, 0.01). Intuitively, it seems that the SCL objective conflicts with CE loss to a certain degree. 

To visualize the relations and interactions between CE loss and SCL loss, we conduct a pilot experiment on the RTE task from GLUE benchmark~\cite{wang2018glue}. We fine-tune two BERT models on RTE. Model A employs only the CE objective for fine-tuning, while model B uses the CE+SCL objective, following the settings from \citet{Gunel2021SupervisedCL}. We run the fine-tuning procedures five times under different random seeds. Model A achieves the accuracy of 65.3 on average (with std. 0.35), and model B achieves the 65.8 on average (with std. 0.84). After adding the SCL objective, model B's performance becomes more unstable than vanilla BERT fine-tuning.

\begin{figure}[tb!]
	\centering
	\includegraphics[width=0.48\textwidth]{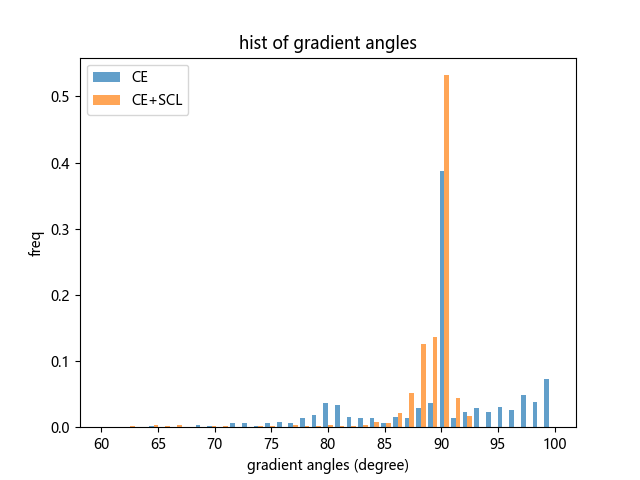}
	\caption{The distribution of gradient angles after the BERT model is fine-tuned on RTE with the CE objective (in blue) or the CE+SCL objective (in orange). }
	\label{fig:gradient_angle}
\end{figure}

To take a closer look into the interactions between CE loss and SCL loss, we separately  compute the gradients derived from the two loss objective, on the fine-tuned model A (or B) with model parameters fixed, $\mathbf{g}_{SCL} = \Delta_{\Theta} \mathcal{L}_{SCL}$ and $\mathbf{g}_{CE} = \Delta_{\Theta} \mathcal{L}_{CE}$. Their angle is given by 
\begin{align}
\cos \gamma & = \dfrac{\mathbf{g}_{SCL} \cdot \mathbf{g}_{CE}}{|\mathbf{g}_{SCL}| \cdot |\mathbf{g}_{CE}|}, \nonumber \\
\gamma & = Argcos(\cos \gamma).
\end{align} 
\noindent The distribution of the gradient angles are plotted in Figure~\ref{fig:gradient_angle}. We can see that, for a BERT fine-tuned solely with CE loss, the gradients' angle $\gamma$ ranges from 75$^{\circ}$ to 105$^{\circ}$, meaning that they have conflicting directions for optimization. The same observation can be made for BERT fined-tuned with the CE+SCL objective. Joint optimization does not make the SCL objective more in alignment with the CE loss in terms of optimization directions. 

In summary, the above observation motivates us to align the two objectives' optimization directions, and as a result, we may obtain better convergence.

\subsection{ACL-Embed}

Now we introduce our first modification to the SCL objective, which is called \underline{a}ligned \underline{c}ontrastive \underline{l}earning via label \underline{embed}ding (ACL-Embed). 

\textbf{Label embedding} \quad The sentence representation $h^{e}_{i}$ given out by the MHA exit will be fed into the classification layer to calculate the logits
\begin{equation}
\mathrm{logits}_i = h^{e}_{i} W_{c},
\end{equation}
where $W \in \mathbf{R}^{d_e \times K}$ is the learn-able parameter matrix of the classifier. Denote the $k$-th column vector of $W$ as $l_{k} \in \mathbf{R}^{d_e}$. The CE loss objective is to maximize the dot product between $h^{e}_{i}$ and $l_{y_i}$ while minimizing the dot products between $h^{e}_{i}$ and the other column vectors of $W$. Thus, one can treat $l_{k}$ as a vector representation of label $k$.  

\textbf{Aligned contrastive loss} \quad Note that SCL~\cite{Gunel2021SupervisedCL} calculates the contrastive learning objective among the samples' representations. With the MHA exit, we consider $h^{e}_{i}$ as the sample representation and feed it into the SCL objective. In contrast, we consider $l_{k}$ ($k \in \mathcal{K}$) as additional $K$ samples $\left\{ (l_k, k), k \in \mathcal{K} \right\}$ to augment the current batch of samples. Thus, we will calculate the new contrastive loss $\mathcal{L}_{acl}$ on the augmented batch of $N+K$ samples via Equation \ref{eq:scl}. We will refer to this objective as the ACL objective. And the overall loss is given by $\mathcal{L} = (1-\lambda) \mathcal{L}_{ce} + \lambda \mathcal{L}_{acl}$.

The idea of ACL-Embed is quite intuitive (depicted in Figure~\ref{fig:framework}). First, the SCL objective is separated from CE loss. Meanwhile, our ACL-Embed objective combines the learning of the classifier with the contrastive loss. In this way, label embeddings serve as anchors to encourage the representations to cluster around them and thus separate away from the other classes. Note that ACL-Embed consists of three parts: (a) contrast among samples, which learns the generic representation across samples; (b) contrast between sample representation and the label embeddings learn to align the representations with the classification task; (c) contrast among the label embeddings, which serves as regularization for the classifier parameters.

\subsection{ACL-Grad}

Although our ACL-Embed method can align the CE objective and the CL-based objective to a certain extent, we still find that, during fine-tuning, the angles between $\mathbf{g}_{ACL}$ and $\mathbf{g}_{CE}$ may be more than 90$^\circ$ in around 20\% of the mini-batches. Thus, we propose a modification to the optimization procedure, called \underline{a}ligned \underline{c}ontrastive \underline{l}earning via \underline{g}radients (ACL-Grad).

% Future work: lambda 根据梯度角度进行调整；
ACL-Grad adaptively modifies the co-efficient $\lambda$ according to the gradient angle. Denote a threshold $\gamma_{thres}$ for gradient angles, then the weight of $\mathcal{L}_{acl}$ on the current batch is given by:
\begin{equation}
	\lambda^{'} = \left\{
	\begin{array}{lr}
		\lambda, & \text{if gradient angle} \ \ \gamma \leq \gamma_{thres}   \\
		0, & \text{otherwise}. 
	\end{array}
	\right.
\end{equation}
Intuitively, the gradient descent direction mainly follows CE loss. ACL-Embed provides an auxiliary regularization to the optimization. On a batch of samples, if the gradient angle is larger than $\gamma_{thres}$, we will discard the gradient information from ACL-Embed to avoid being distracted.  

% Another version of ACL-Grad, called ACL-Grad-v2 can be proposed. ACL-Grad-v2 will scale the gradients from the ACL objective, if the gradient angle is larger than the threshold:

% \begin{equation}
% 	\mathbf{g}_{ACL} = \left\{
% 	\begin{array}{lr}
% 		\mathbf{g}_{ACL} * \min ( \dfrac{\| \mathbf{g}_{CE} \|}{\| \mathbf{g}_{ACL} \|}, 0.5 ), & \text{if} \ \ \gamma > \gamma_{thres}   \\
% 		\mathbf{g}_{ACL}, & \text{otherwise} 
% 	\end{array}
% 	\right.
% \end{equation}

\subsection{Cross-layer ACL as cross-sample knowledge distillation}

This work mainly adopts the 2ST training strategy for multi-exit BERT fine-tuning (in Section \ref{sec:multi_exit_bert_training}). After the BERT backbone and the last layer's exit is fine-tuned, the intermediate exits will distill knowledge from the last exit to gain better performances. However, we believe that the usual knowledge distillation method like~\cite{Hinton2015DistillingTK}, feature distillation, or attention matrix distillation~\cite{jiao2019tinybert} does not provide knowledge of the semantic relations or comparisons among different samples. In addition, the intermediate classifiers do not receive direct guidance from the last exits. The above analysis encourages us to propose a novel knowledge distillation method to discover the potentials of intermediate layers better. 

Assume we are now fine-tuning the $m$-th layer's exit of BERT, which represent the batch of samples into vectors $h_{i}^{e, m}$ ($i \in \mathcal{I}$). Its classification layer has label embedding $l_k^{m}$. Despite the parameters are fixed, the last exit $M$ can provide representations $h_{i}^{e, M}$ ($i \in \mathcal{I}$) for the batch of samples and the label embeddings $l_k^{M}$ ($k \in \mathcal{K}$). We propose a novel cross-layer contrastive learning objective, ACL-CL. ACL-CL calculates the contrastive loss using Equation \ref{eq:scl} on the augmented batch of $2N + 2K$ representations, including $h_{i}^{e, m}$, $l_k^{m}$, $h_{i}^{e, M}$ and $l_k^{M}$. During the training of an intermediate exit, ACL-CL can also incorporate ACL-Grad method to make the optimization smoother. 

Intuitively, ACL-CL can be beneficial due to following three reasons. 
\begin{itemize}
    \item First, ACL-CL objective includes the ACL-Embed objective on the intermediate layer.
    \item Second, the shallow exit can directly learn from the label embeddings from the last exit, thus improving its representation quality.
    \item Third, the shallow layer of BERT is less capable of learning a good representation. However, with our ACL-CL objective, it can learn generic features among samples from the deeper layer.  
\end{itemize}   

ACL-CL can be seen as an effective knowledge distillation method for the intermediate layers, which does not only extract knowledge sample by sample but also distill knowledge on how the samples are contrasted and how to form better label embeddings for classification.

\section{Experiments}

\subsection{Datasets}

We evaluate our proposed approach to the classification tasks on GLUE benchmark~\cite{wang2018glue}. We only exclude the STS-B task since it is a regression task, and we exclude the WNLI task following previous work \citep{devlin2018bert,Jiao2020TinyBERTDB,Xu2020BERTofTheseusCB}.

\subsection{Baseline methods}

In the normal BERT fine-tuning scenario, we compare ourselves with two strong baselines: (1) fine-tuning BERT/RoBERTa with CE loss~\cite{devlin2018bert}; (2) fine-tuning with CE+SCL objective~\cite{Gunel2021SupervisedCL}. 

In order to demonstrate our ACL's effectiveness on multi-exit BERT training and inference speedup, we compare with the following types of baselines: 

\textbf{Directly training of intermediate layers} We experiment with the naive baseline that directly utilizes the first 6 layers of the original BERT/RoBERTa with an MHA exit on the top, denoted by BERT-6L/RoBERTa-6L respectively. 

\textbf{Model compression methods} \quad For model parameter pruning, we include the results of LayerDrop \citep{Fan2020ReducingTD} and attention head pruning \citep{Michel2019AreSH}. For knowledge distillation based pre-training, we include DistillBERT \citep{Sanh2019DistilBERTAD} and TinyBERT~\cite{Jiao2020TinyBERTDB}. For module replacing, we include BERT-of-Theseus \citep{Xu2020BERTofTheseusCB}.

\textbf{Existing training methods for multi-exit BERT} \quad We compare our ACL framework with three representative multi-exit BERT training methods: (a) JT~\cite{Zhou2020PABEE,Teerapittayanon2016BranchyNetFI}; (b) 2ST~\cite{xin2020deebert,liu2020fastbert}; (c) ALT~\cite{xin2021berxit}. We implement 2ST with both the linear exit and the MHA exit.

% \textbf{Dynamic early exiting methods} \quad To verify that ACL-BERT can enhance the performances of dynamic exiting methods, we adopt the early exiting strategies of BranchyNet (xxx, ) and PABEE (xxxx, Zhou et al., 2020) on BERT and will report the speedup-performance curves. 

\subsection{Experimental settings}

\textbf{Devices} \quad We implement our ACL framework on the base of HuggingFace's Transformers. We conduct our experiments on Nvidia V100 16GB GPUs.

\textbf{Backbone models} \quad All of the experiments are built upon the Google BERT base~\cite{devlin2018bert} or RoBERTa base~\cite{Liu2019RoBERTaAR} models. 

% We ensure fair comparison by setting the hyper-parameters related to the PLM backbones the same with HuggingFace Transformers~\cite{Wolf2019HuggingFacesTS}.

\textbf{Hyper-parameter settings} \quad The hidden dimension $d_e$ in the MHA is set to 64 so that the extra parameters introduced by MHA exits will be less than 0.2\% of the pre-trained models. We use the AdamW optimizer~\cite{Loshchilov2019DecoupledWD} to finetune the pre-trained models with a 0.01 weight decay. We train each backbone model and intermediate exits for 15 epochs and use a linear learning rate decay from 2e-5 to 0 with 10\% of the total optimization steps as warm-up. We set the dropout rate to 0.1 for all layers. Batch size and gradient accumulation are used so that there are around two hundred gradient update steps per epoch. The temperature factor $\tau$ is chosen as 0.5, the coefficient $\lambda$ is set at 0.02, and the gradient angle threshold is set to 90$^{\circ}$. We mainly adopt the 2ST training method with MHA exits for multi-exit BERT/RoBERTa training. Other hyper-parameters follow the Huggingface Transformers~\cite{Wolf2019HuggingFacesTS}.

\subsection{Overall Comparison}

\begin{table*}[tb!]
\centering

\resizebox{0.99\textwidth}{!}{
\begin{tabular}{cccccccccc}
\hline

Model & Method & \#Flops & CoLA & MNLI & MRPC  & QNLI  & QQP & RTE & SST-2  \\ 

\hline
\multicolumn{10}{c}{\emph{Full model}} \\
\hline 

BERT &  CE   & 6.6G  &  57.3 &  82.9/83.4  & \textbf{89.6}  &  89.5 & 88.7   & 65.3 &  \textbf{92.8} \\ 

  &  CE+SCL   & 6.6G  &  57.9  &   82.6/82.7  &   88.7  &  89.7  &  88.5   &  65.8   &  92.1 \\
  % &  CE+SSCL   & 6.6G  &   57.8  &    82.8/83.6  &  89.2  & 89.4  &   88.2   & 65.6  &  92.3 \\
  &  ACL   &  6.6G  &  \textbf{58.5}  &    \textbf{83.0/83.7}  &   89.5  &   89.7  &   \textbf{88.9}   &  \textbf{66.2}   &  92.7 \\ 
\hline 

RoBERTa &  CE   & 6.7G  &  62.4 &  86.9/86.5  & 90.6  &  91.8  & \textbf{90.3}   & 73.6 &  93.7 \\ 
  &  CE+SCL   & 6.7G  &   62.7   &  \textbf{87.1/86.7}  &  90.7  &  92.1   &  89.5  &   74.2   &  93.6   \\
 % &  CE+SSCL   & 6.7G  &  62.6   &  86.8/86.7  &   90.5   &  91.6   & 90.0   & 73.8  &   93.3 \\
  &  ACL   &  6.7G  &   \textbf{63.1}   &    86.8/86.9  &  \textbf{90.8}    &   \textbf{92.3}  &  90.2   &  \textbf{75.3}    &   \textbf{94.2}  \\ 
\hline

\hline
\multicolumn{10}{c}{\emph{Efficient models}} \\
\hline

BERT-6L &  -  &   3.3G  & 43.2  & 81.4/81.6  &  84.9 &  87.9  &  87.2 &   61.0   &  90.3 \\
RoBERTa-6L &  -  &  3.4G  &  45.6   &  82.5/83.9  &  86.9   & 89.8  &  88.6  &  67.8  &  91.6 \\ 

LayerDrop-6L &  -  &  3.3G  &  43.7  &  82.3/82.4  &  86.5   &  89.3  &  89.1   &  67.3   &   91.8   \\
HeadPrune &  -  & 4.7G  &   45.7 &   78.0/79.7  &  80.2  &   86.1  &  84.7  &   64.5   &   91.0   \\

DistilBERT (6L) &  -  &  3.3G  & 46.1 &   82.1/82.0   &    86.2  &  88.3   &  88.5  &      64.1    &   91.3    \\
TinyBERT (6L) &  -  &  3.3G  &  \textbf{46.3}  &  \textbf{83.6/83.8}  &   87.2  &  90.3  &  \textbf{89.1}  &       68.6   &  91.9  \\

BERT-of-Theseus &  -  &  3.3G  &  44.7  &  81.4/81.9  &  85.3  &  88.1  &  88.6    &      66.2   &   91.5  \\

\hline

ACL-BERT-6L  &  ACL-CL  &  3.3G  &   44.7   &  82.0/82.1  &  85.1  &   88.5   &  87.7  &  63.6   &  90.8 \\
ACL-RoBERTa-6L   &  ACL-CL  &  3.3G  &  46.2  &   83.3/83.5   &  \textbf{87.5}   &   \textbf{90.5}  &   88.9   &   \textbf{69.5}   &   \textbf{92.3}  \\

\hline

\end{tabular}}

\caption{\label{tab:main_results} Our ACL framework and baselines' performance on the GLUE tasks. In this table, we mainly compare the performances of ACL on the full BERT/RoBERTa or the 6-th layer against baselines. Performance metrics for each task strictly follow the settings of the GLUE benchmark. }

\end{table*}

Table \ref{tab:main_results} reports the main results on the GLUE benchmark with BERT and RoBERTa as the backbone models. The last two lines of the table are the 6-th layer's performances of the multi-exit BERT/RoBERTa enhanced by our ACL framework. All baseline models are run with the original authors' open-sourced codes.

\textbf{Performance comparisons for BERT fine-tuning} \quad From Table \ref{tab:main_results}, we can see that by adding our ACL objective, BERT obtains performance improvements against vanilla fine-tuning or CE+SCL on 5 of the 7 GLUE tasks. RoBERTa also benefits from the ACL objective. 

\textbf{Performance comparisons of efficient models} \quad Table \ref{tab:main_results} also reports the comparison among model inference speedup methods. With the help of our ACL-CL objective, the intermediate exits of BERT can learn better semantic representations for the samples and the labels. As a result, ACL-BERT-6L can outperform BERT-6L. With a better backbone, ACL-RoBERTa-6L can outperform or perform comparably with the SOTA static BERT compression methods like TinyBERT. Note that TinyBERT requires that the compressed model be further pre-trained, while our method does not. 

\subsection{Comparisons against multi-exit BERT training methods} 
\begin{figure*}[h]
\centering
	\subfigure[MRPC]{%
		\includegraphics[width=0.33\textwidth]{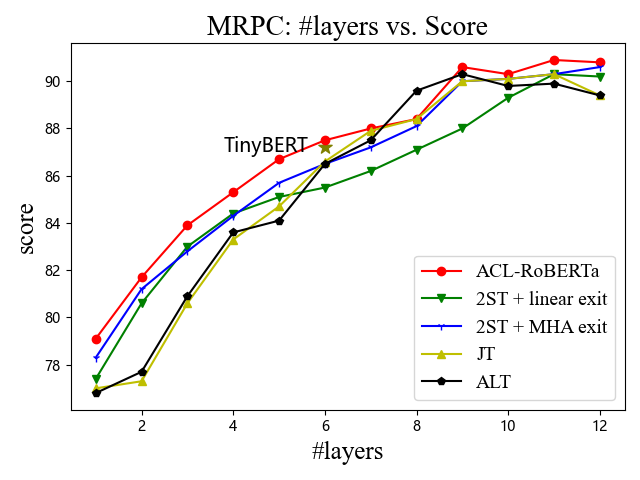}
		\label{subfig:mrpc_layer2score_roberta}
	}\hspace{-3mm}
	\subfigure[RTE]{%
		\includegraphics[width=0.33\textwidth]{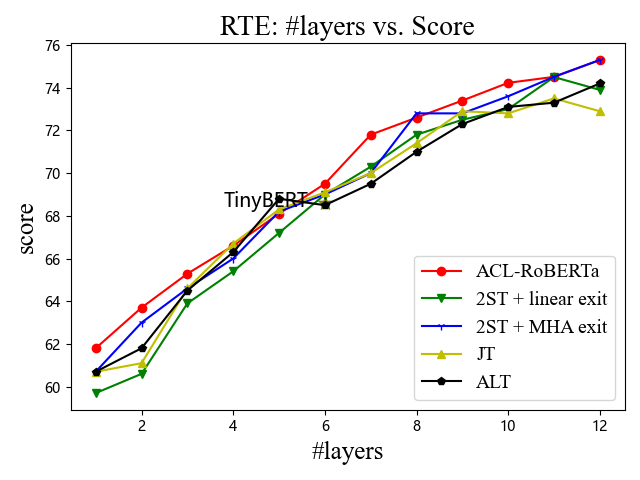}
		\label{subfig:rte_layer2score_roberta}
	}\hspace{-3mm}
	\subfigure[SST-2]{%
		\includegraphics[width=0.33\textwidth]{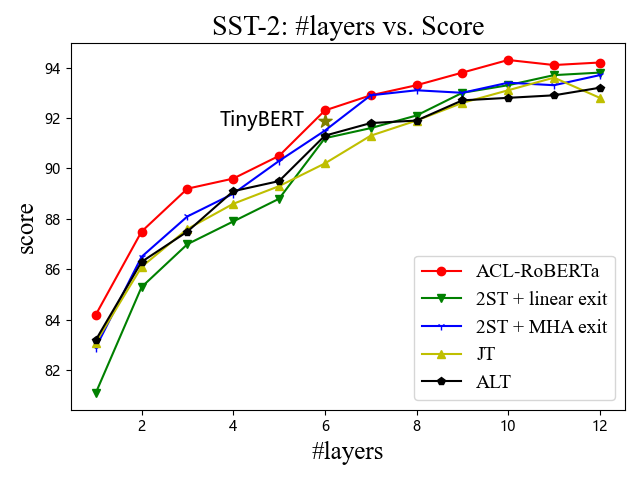}
		\label{subfig:sst-2_layer2score_roberta}
	}\hspace{-3mm}
\caption{\label{fig:layer_vs_scores}The layer-score curves for different multi-exit BERT training methods, with the RoBERTa backbone. TinyBERT is represented as a dot since it is a static compression method.  }
	
\end{figure*}

\quad To further demonstrate our method's effectiveness in improving the intermediate exits, we compare our ACL-RoBERTa with the previous SOTA multi-exit BERT training methods. For visualization, we plot the curve between the layer depth and the performance score (layer-score curve) in Figure \ref{fig:layer_vs_scores}.\footnote{Due to the length limit, layer-score curves for other GLUE tasks can be found in Figure \ref{fig:layer_vs_scores_appendix} of Appendix \ref{sec:layer_score_curves}. } The results show that our ACL-RoBERTa outperforms the baseline multi-exit BERT training methods on most of the layers. Note that the performance margins on the shallow exits are more significant than those on the deeper exits, showing that our ACL framework effectively distills knowledge from the deeper exit to guide the shallow ones.

In addition, from figure \ref{fig:layer_vs_scores}, we can clearly see that MHA exit significantly improve the performance of 2ST training with a simple linear head. 

To show that our ACL framework's performance gains in multi-exit BERT training do not rely on the specific backbone models, we also plot the layer-score curves on the RTE and MRPC tasks with the BERT-base backbone (Figure \ref{fig:layer_vs_scores_bert} in Appendix \ref{sec:layer_score_curves}). The results show that ACL-BERT can also outperform the baseline multi-exit BERT training methods.

% For better visualization of the results, we plot the curve between the layer depth in BERT and the performance score (layer-score curve) in Figure \ref{layer_vs_scores}. The performances of TinyBERT and DistillBERT are represented by dots in these figures, since they are static compression methods. 

%  From Figure \ref{fig:layer_vs_scores}, we can see that ACL-RoBERTa outperforms the existing multi-exit model training methods on most of the RoBERTa layers. Similar observations can also be made on the BERT backbone. Note that the performance margins on the shallow exits are more significant than those on the deep exits, showing that our model is effective in improving the shallow early exits' performances. From the figure \ref{}, we can clearly see that MHA exit significantly improve the performance of 2ST training with a simple linear head. 

% \subsection{Ablation on the backbone models for MEA training}

% Figure \ref{} demosntrates the effectiveness of ACL in the multi-exit RoBERTa training. However, our ACL framework is off-the-shelf and can be applied to other backbone models without effort. Figure \ref{} plots the layer-score curves of different methods on the MRPC task, with BERT and ALBERT as backbone models. The results in Figure \ref{} clearly show the superiority of ACL-BERT and ACL-ALBERT against the existing MEA training method. 

\subsection{Ablation on the ACL framework}

In order to analyze which components of our ACL framework are beneficial for multi-exit BERT fine-tuning, we step by step reduce our whole ACL framework into the vanilla 2ST training: (a) first, not use ACL objective during the backbone fine-tuning in the first stage; (b) second, reduce the ACL-CL to ACL-Embed on the intermediate exits, so the last exit will only provide logits as soft targets; (c) third, further exclude the ACL-Grad module for intermediate exits' training; (d) fourth, substitute the ACL-Embed objective into CE+SCL; (e) fifth, drop the SCL objective. The ablation study is done on MRPC, RTE, and SST-2 tasks, with RoBERTa backbone. 

% To further demonstrate our ACL framework's effectiveness in multi-exit BERT fine-tuning, we step by step reduce our whole cross-layer ACL framework to the vanilla 2ST training framework: (a) since ACL can improve the final exit's performances and as a result improve the intermediate layers' score by providing better guidance via ACL objective, we now ask the backbone and final exit be fine-tuned with CE objective; (b) during the cross-layer ACL learning, label embeddings from the teacher exits are not included; (c) further exclude the involvements of teachers' representations; (d) further discard ACL-Grad during the training of intermediate exits; (e) now the ACL-Embed objective be dropped, and the intermediate layers will only optimize with the soft CE objective. The ablation study is done on the RTE, MRPC and SST-2 tasks. 

Table \ref{tab:ablation_on_acl} reports the average performance scores across all the layers of the RoBERTa backbone. The results show that each component in our whole ACL framework is beneficial for the multi-exit BERT training procedure. Our ACL objectives can effectively guide the optimization of weak intermediate layers and result in better performances.

\begin{table}[tb!]
\centering
\resizebox{0.46\textwidth}{!}{
\begin{tabular}{c|ccc}
\hline
Method &  \multicolumn{3}{c}{cross-layer average score}  \\ 
- & MRPC &  RTE  &  SST-2   \\
\hline
% \multicolumn{4}{l}{\emph{ALBERT}} \\
% \hline 
ACL-RoBERTa &   86.9   &   69.8   &  91.3  \\ 
- CE backbone  &  86.9  &   69.7  &   91.1    \\ 
- w/o ACL-CL  &  86.7  &   69.4   &  91.0 \\ 
- w/o ACL-Grad  & 86.5   &  69.3  &  90.8 \\
- CE+SCL  &  86.4  &  69.3  &  90.5 \\
- 2ST &  86.3 &  69.2   &  90.6  \\ 
\hline
\end{tabular}}
\caption{\label{tab:ablation_on_acl} Ablation study on the components of our ACL method. Cross-layer average performance scores are reported. }
\end{table}

\subsection{Visualization of the intermediate exits' representations}

To demonstrate how our ACL objective improves the semantic representations, we draw the t-SNE plots of the semantic representations on the 3-th exit, with CE+SCL objective (in Figure \ref{subfig:layer_3_tsne_1} in Appendix \ref{sec:t_sne_plot}) or with our ACL method (in Figure \ref{subfig:layer_3_tsne_2} in Appendix \ref{sec:t_sne_plot}). The samples are from the dev set of MRPC. With CE+SCL objective, the 3-th exit achieves the performance score of 82.8\%, while the score is 83.6\% under the ACL method. As is shown in Figure \ref{subfig:layer_3_tsne_1} and \ref{subfig:layer_3_tsne_2}, our ACL help the model to learn more discriminative representations for the input samples, and the label embeddings learned with ACL are in alignment with the input samples from the corresponding classes.  

% on the 4-th exit of BERT fine-tuned on the MRPC task, with CE+SCL objective (in Figure \ref{}) and with our ACL method (in Figure \ref{}). With CE+SCL objective, the 4-th exit achieves the performance score of 82.8\%, while the score is 83.6\% under the ACL method. As is shown in Figure \ref{} and \ref{}, our ACL help the model to learn more discriminative representations for the input samples, and the label embeddings learned with ACL are in alignment with the input samples from the corresponding classes. 

\subsection{Parameter sensitivity of the $\lambda$ coefficient} 

We now look into how the value of co-efficient $\lambda$ affects the multi-exit BERT training by setting $\lambda$ to be one of \{ 0.005, 0.01, 0.02, 0.1, 0.5, 1.0\} for both the CE+SCL method and our ACL method. The cross-layer average scores are reported in Table \ref{tab:parameter_sensitivity_study} in the Appendix \ref{sec:parameter_sensitivity}. The results show that our ACL framework is not sensitive to the values of $\lambda$, whereas the CE+SCL method is prone to be affected by $\lambda$. This observation shows that our ACL method can conveniently apply to other tasks.

\section{Conclusion}

In this work, we take a close look at the supervised contrastive learning objective~\cite{Gunel2021SupervisedCL} and find that the CE objective and SCL objective often conflict with each other, thus hindering better optimization. To overcome this issue, we propose the \underline{A}ligned \underline{C}ontrastive \underline{L}earning (ACL) framework. ACL consists of three main components: (a) ACL-Embed, which aligns the label embeddings from the classifier with the samples' representations; (b) ACL-Grad, which selectively shuts down the ACL-Embed loss term if gradients from ACL-Embed and CE have very different directions; (c) ACL-CL, which performs cross-layer contrastive learning at the shallow exit to extract knowledge from the deeper exit. Extensive experiments on the GLUE benchmark results demonstrate that: (a) our ACL framework outperforms the CE+SCL approach in BERT fine-tuning; (b) ACL-CL outperforms the existing SOTA multi-exit BERT training method.

\section*{Limitation}

% \noindent 虽然META-BERT能够进一步提升Multi-exit模型的训练效果，但是我们认为其仍然存在一些限制。我们通过adpater对主干模型进行微调的方式将会增加模型可训练的参数量。在未来的工作中我们将探索更高效的模型微调方法。其次，在本文中我们只关注了分类任务，有各种其他的任务需要探索，不同的任务上将会有不同的挑战。

Although our ACL framework is shown to be effective in improving the multi-exit BERT training, it still has certain limitations that need to be addressed in the future: (a) In this work, we demonstrate our framework's performances on the sentence classification tasks or sentence pair classification tasks. In future works, we would like to extend our work to broader tasks such as sequence labeling, relation extraction, and dependency parsing. (b) Our ACL framework effectively improves the training of multi-exit BERT, which suggests its value in the study of BERT inference speedup methods. We would like to investigate whether our ACL can extend to static model compression methods like the TinyBERT framework.

\section*{Ethics Statement}
% \section{Broader impact and ethical consideration}

Our ACL-BERT is designated to improve the fine-tuning of BERT models and the training of multi-exit BERT, providing a better backbone model for dynamic early exiting. Our work could help make the applications of natural language processing more accessible by reducing computational costs. In addition, this work can be seen as an effort to reduce BERT-based models' carbon footprints. Furthermore, It does not introduce new ethical concerns.

\bibliography{custom}
\bibliographystyle{acl_natbib}

\appendix

\section{Additional related works}
\label{sec:appendix_additional_related_works}

\textbf{Contrastive learning in the unsupervised setting} \quad Contrastive learning was first proposed by~\citet{Hadsell2006DimensionalityRB}. Since the seminal work of InfoNCE~\cite{Oord2018RepresentationLW}, contrastive learning has been widely applied in the self-supervised learning settings~\cite{Wu2018UnsupervisedFL,Oord2018RepresentationLW,Tian2020ContrastiveMC,He2020MomentumCF,Chen2021ExploringSS}. MoCo~\cite{He2020MomentumCF} proposes a dictionary to maintain a negative sample set, thus increasing the number of negative sample pairs. \citet{Chen2021ExploringSS} construct a positive-only contrastive learning framework, by using a stop-gradient operation to prevent collapsing solutions. Contrastive learning is also widely applied in NLP. CERT~\cite{Fang2020CERTCS} use back-translations to generate positive views of an sentence and apply the MoCo framework for contrastive learning based pre-training. SimCSE~\cite{Gao2021SimCSESC} propose to use two separate forward passes with dropout to generate positive views and apply CL to unsupervised sentence representations. 

\section{Additional preliminaries}
\label{sec:additional_preliminaries}

\subsection{Contrastive objective in the unsupervised setting}
\label{subsec:appendix_preliminaries_cl}

The effectiveness of contrastive learning is widely confirmed in the self-supervised settings~\cite{Tian2020ContrastiveMC,He2020MomentumCF}. Here we give a brief introduction of the self-supervised InfoNCE~\cite{Oord2018RepresentationLW} objective widely used in the contrastive learning literature. Given $N$ training samples $\{x_{i}\}_{i=1}^{N}$ with a number of augmented samples, where each sample has at least one augmented sample in the dataset. Let $j(i)$ be the index of the augmented one derived from the $i$-th sample, the standard contrastive loss is defined as:
\begin{equation}
    \mathcal{L}_{self} = \dfrac{1}{N} \sum_{i\in \mathcal{I}} - \log \dfrac{ \exp (z_{i} \cdot z_{j(i)} / \tau ) }{ \sum_{a \in \mathcal{A}_{i} } \exp (z_{i} \cdot z_{a} / \tau ) },
\end{equation}
where $z_{i}$ is the normalized representation of $x_{i}$, $\mathcal{A}_{i} \coloneqq \mathcal{I}\ \{i\}$ is the set of indexes of the contrastive samples, the $\cdot$ operation denotes the dot product and $\tau \in \mathbf{R}^{+}$ is the temperature factor. Here, contrastive learning treats the $i$-th sample as an anchor, the $j(i)$-th sample is a positive sample and the remaining $2N-2$ samples are negative samples regarding to the $i$-th sample.

\subsection{Introduction to the multi-exit training methods}
\label{subsec:mutli_exit_training}

\noindent Here we introduce two training methods of multi-exit BERT: JT and ALT.

\textbf{JT}. Perhaps the most straightforward fine-tuning strategy is to minimize the sum of all classifiers’ loss functions and jointly update all parameters in the process.  We refer to this strategy as JT.  The loss function is:
\begin{equation}
    \mathcal{L}_{JT} = \sum_{m=1}^M \mathcal{L}_m^{CE}
    \label{training:joint}
\end{equation}
\noindent where $\mathcal{L}_m^{CE}=\mathcal{L}_m^{CE}(y,f_m(x;\theta_m))$ denotes the cross-entropy loss of the m-th exit. This method is adopted by \cite{Teerapittayanon2016BranchyNetFI, Kaya2019ShallowDeepNU, Zhou2020PABEE, Zhu2021LeeBERTLE}

\textbf{ALT}. It alternates between two objectives (taken from Equation \ref{training:joint} and \ref{training:2st}) across different epochs, and it was proposed by BERxiT \cite{xin2021berxit}:

\begin{flalign}
    & \mathrm{Odd}:\mathcal{L}_{stage1} = \mathcal{L}^{CE}_{M}(y_i, f_M(x_i;\theta_M))\label{training:odd} \\
    & \mathrm{Even}:\mathcal{L}_{joint} = \sum_{m=1}^M \mathcal{L}_m^{CE}
    \label{training:even}
\end{flalign}

\section{Layer-score curves for GLUE tasks}
\label{sec:layer_score_curves}

\begin{figure*}[h]
	\centering
	\subfigure[CoLA]{%
		\includegraphics[width=0.46\textwidth]{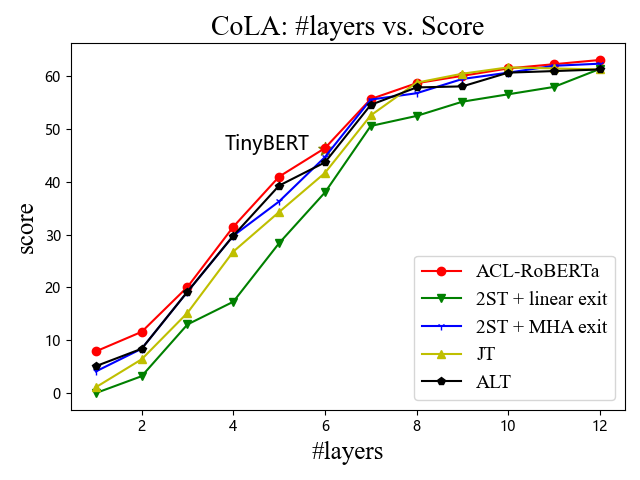}
		\label{subfig:cola_layer2score_roberta}
	}
	\subfigure[QNLI]{%
		\includegraphics[width=0.46\textwidth]{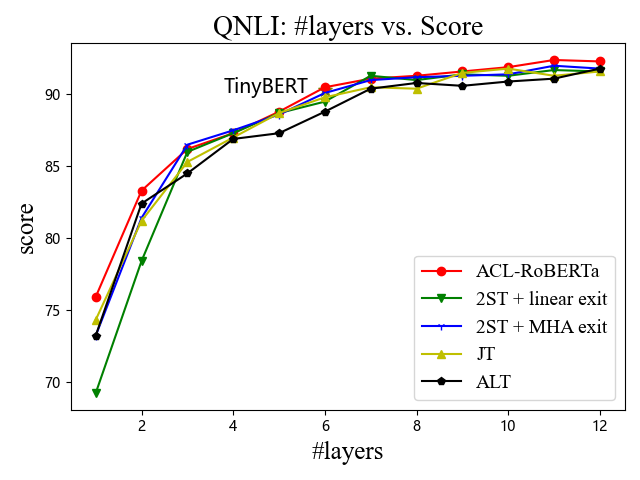}
		\label{subfig:qnli_layer2score_roberta}
	}
	\subfigure[QQP]{%
		\includegraphics[width=0.46\textwidth]{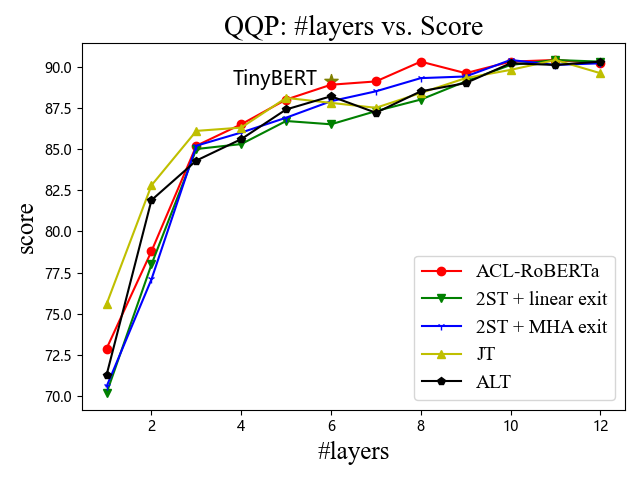}
		\label{subfig:qqp_layer2score_roberta}
	}
	\subfigure[MNLI]{%
		\includegraphics[width=0.46\textwidth]{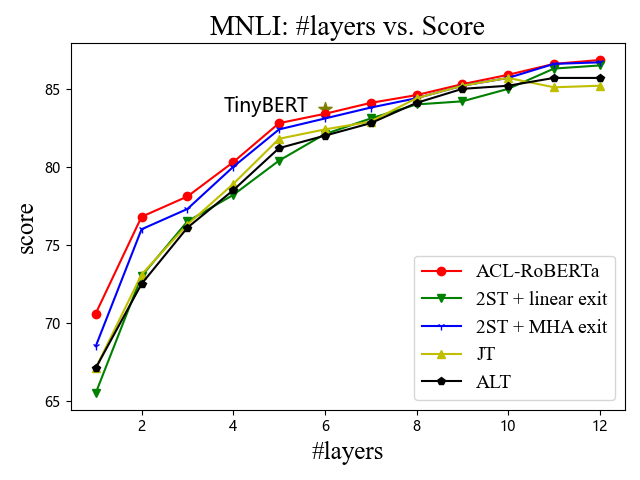}
		\label{subfig:mnli_layer2score_roberta}
	}
	\caption{\label{fig:layer_vs_scores_appendix}The layer-score curves for different multi-exit BERT training methods, with the RoBERTa backbone. The $x$-axis is the depth of the exit (or the number of layers before entering this exit), the $y$-axis is the performance metrics following GLUE \citep{Wang2018GLUEAM}. The performance of TinyBERT is represented by a dot since it is a static model compression method.   }
\end{figure*}

In Figure \ref{fig:layer_vs_scores}, we present the layer-score curves for 3 of the GLUE tasks, to compare MEAL-BERT with other baseline models. Here, we present layer-score curves on the rest 4 tasks (CoLA, QNLI, QQP, MNLI) in Figure \ref{fig:layer_vs_scores_appendix}. 

\begin{figure*}[h]
	\centering
	\subfigure[MRPC]{%
		\includegraphics[width=0.46\textwidth]{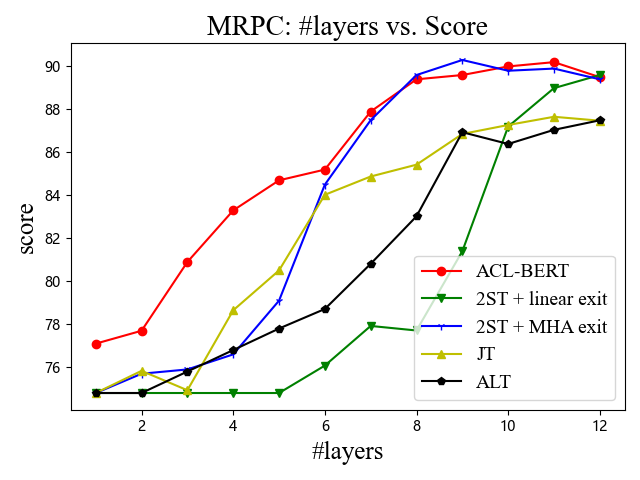}
		\label{subfig:mrpc_layer2score_bert}
	}
	\subfigure[RTE]{%
		\includegraphics[width=0.46\textwidth]{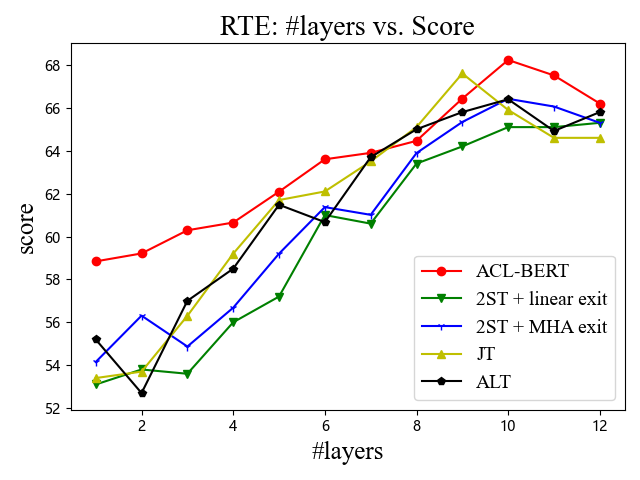}
		\label{subfig:rte_layer2score_bert}
	}
	\caption{\label{fig:layer_vs_scores_bert}The layer-score curves for different multi-exit BERT training methods, with the BERT-base backbone.   }
\end{figure*}

In addition, we present the layer-score curves when the BERT-base is the backbone model, in Figure \ref{fig:layer_vs_scores_bert}. The results show that our ACL framework also outperform other baseline multi-exit BERT training method with BERT base as the backbone.

\section{The t-SNE plots on the third exit}
\label{sec:t_sne_plot}

\begin{figure*}[h]
	
	\centering
	\subfigure[CE+SCL]{%
		\includegraphics[width=0.43\textwidth]{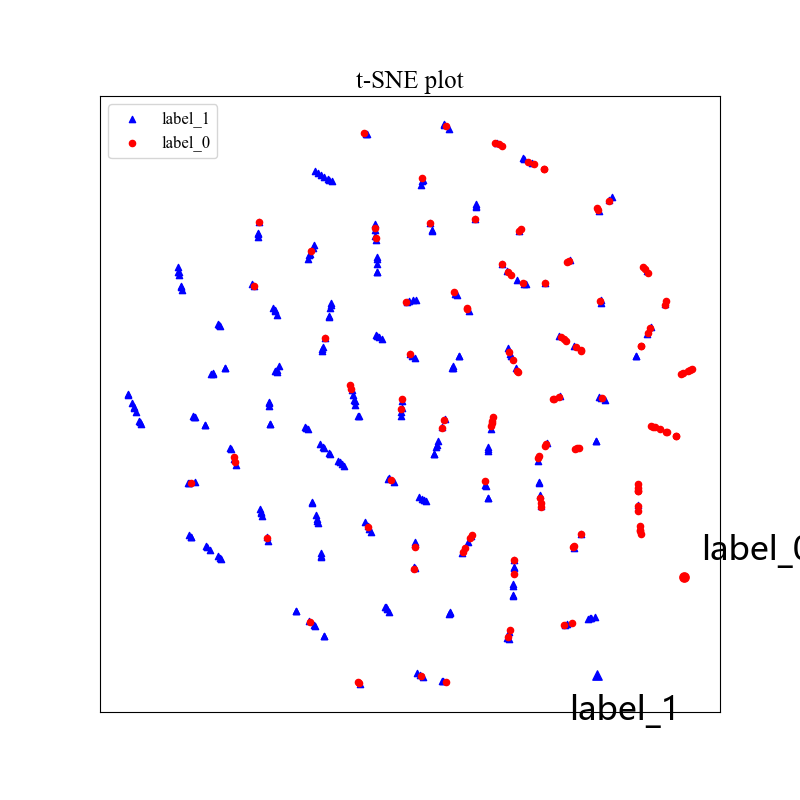}
		\label{subfig:layer_3_tsne_1}}
	\subfigure[ACL]{%
		\includegraphics[width=0.43\textwidth]{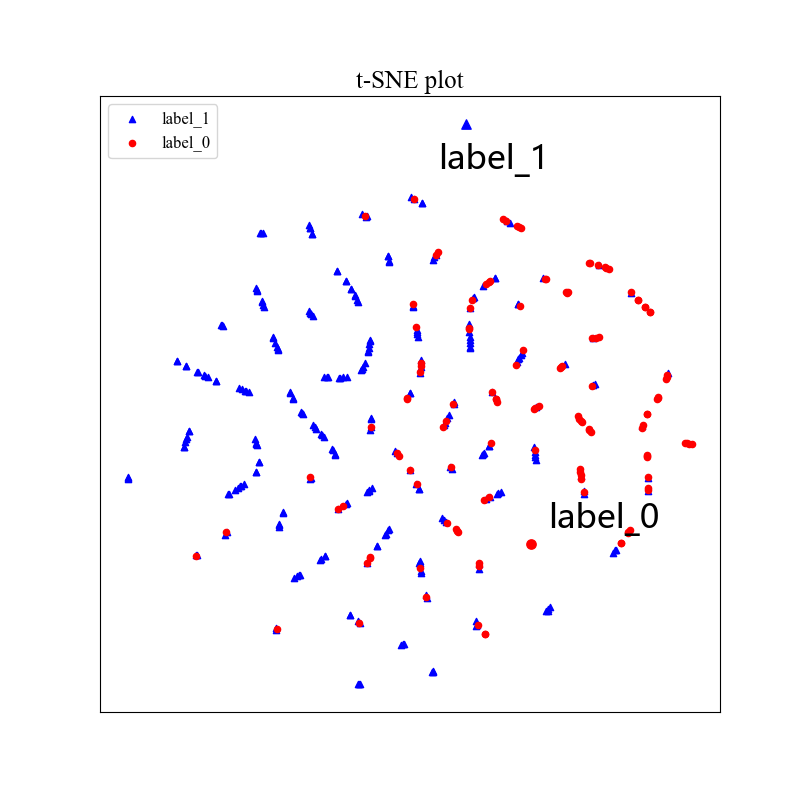}
		\label{subfig:layer_3_tsne_2}}
	\caption{\label{fig:layer_3_tsne}The t-SNE plots of the semantic representations on the 3-th exit of RoBERTa fine-tuned on the MRPC task, with (a) the CE+SCL objective, or (b) our ACL objective. }
	
\end{figure*}

The t-SNE plots of the semantic representations on the 3-th exit, with CE+SCL objective (in Figure \ref{subfig:layer_3_tsne_1}) or with our ACL method (in Figure \ref{subfig:layer_3_tsne_2}). 

\begin{table}[tb!]
\centering
\resizebox{0.3\textwidth}{!}{
\begin{tabular}{c|cc}
\hline
$\lambda$  &  CE+SCL  &   ACL   \\
0.005  &   69.1   &  69.7 \\ 
0.01   &  69.3   &   69.8  \\
0.02  &  69.3   &   69.8   \\
0.1  &  69.0   &   69.7   \\
0.5  &   68.8  &    69.7  \\
1.0  &   68.6  &   69.6  \\
\hline

\end{tabular}}
\caption{\label{tab:parameter_sensitivity_study}The parameter sensitivity study of the co-efficient $\lambda$ on the RTE task. Cross-layer average scores are reported. }

\end{table}

\section{Results for the parameter sensitivity study}
\label{sec:parameter_sensitivity}

In Table \ref{tab:parameter_sensitivity_study}, we will report the cross-layer average performance scores, when the weight $\lambda$ varies for the CE+SCL method, or our ACL method. The backbone model is RoBERTa, and we use the RTE task for this parameter sensitivity study.

\end{document}